\documentclass[sigconf]{acmart}

\usepackage{hyperref}
\usepackage{url}
\usepackage{graphicx}
\usepackage{multirow}
\usepackage{booktabs}
\usepackage{makecell}
\usepackage{graphicx}
\usepackage{subcaption}
\usepackage{tablefootnote}

\AtBeginDocument{%
  \providecommand\BibTeX{{%
    \normalfont B\kern-0.5em{\scshape i\kern-0.25em b}\kern-0.8em\TeX}}}

\copyrightyear{2020} 
\acmYear{2020} 
\setcopyright{acmcopyright}\acmConference[KDD '20]{Proceedings of the 26th ACM SIGKDD Conference on Knowledge Discovery and Data Mining}{August 23--27, 2020}{Virtual Event, CA, USA}
\acmBooktitle{Proceedings of the 26th ACM SIGKDD Conference on Knowledge Discovery and Data Mining (KDD '20), August 23--27, 2020, Virtual Event, CA, USA}
\acmPrice{15.00}
\acmDOI{10.1145/3394486.3403172}
\acmISBN{978-1-4503-7998-4/20/08}

\begin{document}
\fancyhead{}
\title{LayoutLM: Pre-training of Text and Layout for \\ Document Image Understanding}

\author{Yiheng Xu}
\email{charlesyihengxu@gmail.com}
\authornote{Equal contributions during internship at Microsoft Research Asia.}
\affiliation{
  \institution{Harbin Institute of Technology}
}

\author{Minghao Li}
\email{liminghao1630@buaa.edu.cn}
\authornotemark[1]
\affiliation{
  \institution{Beihang University}
}

\author{Lei Cui}
\email{lecu@microsoft.com}
\affiliation{
  \institution{Microsoft Research Asia}
}

\author{Shaohan Huang}
\email{shaohanh@microsoft.com}
\affiliation{
  \institution{Microsoft Research Asia}
}

\author{Furu Wei}
\email{fuwei@microsoft.com}
\affiliation{
  \institution{Microsoft Research Asia}
}

\author{Ming Zhou}
\email{mingzhou@microsoft.com}
\affiliation{
  \institution{Microsoft Research Asia}
}


\begin{abstract}
  Pre-training techniques have been verified successfully in a variety of NLP tasks in recent years. Despite the widespread use of pre-training models for NLP applications, they almost exclusively focus on text-level manipulation, while neglecting layout and style information that is vital for document image understanding. In this paper, we propose the \textbf{LayoutLM} to jointly model interactions between text and layout information across scanned document images, which is beneficial for a great number of real-world document image understanding tasks such as information extraction from scanned documents. Furthermore, we also leverage image features to incorporate words' visual information into LayoutLM. To the best of our knowledge, this is the first time that text and layout are jointly learned in a single framework for document-level pre-training. It achieves new state-of-the-art results in several downstream tasks, including form understanding (from 70.72 to 79.27), receipt understanding (from 94.02 to 95.24) and document image classification (from 93.07 to 94.42). The code and pre-trained LayoutLM models are publicly available at \url{https://aka.ms/layoutlm}.
\end{abstract}

\ccsdesc[500]{Information systems~Business intelligence}
\ccsdesc[500]{Computing methodologies~Information extraction}
\ccsdesc[500]{Computing methodologies~Transfer learning}
\ccsdesc[500]{Applied computing~Document analysis}

\keywords{LayoutLM; pre-trained models; document image understanding}


\maketitle

\section{Introduction}

\begin{figure*}[t]
\centering
    \begin{subfigure}[b]{0.25\textwidth}
        \includegraphics[width=\textwidth]{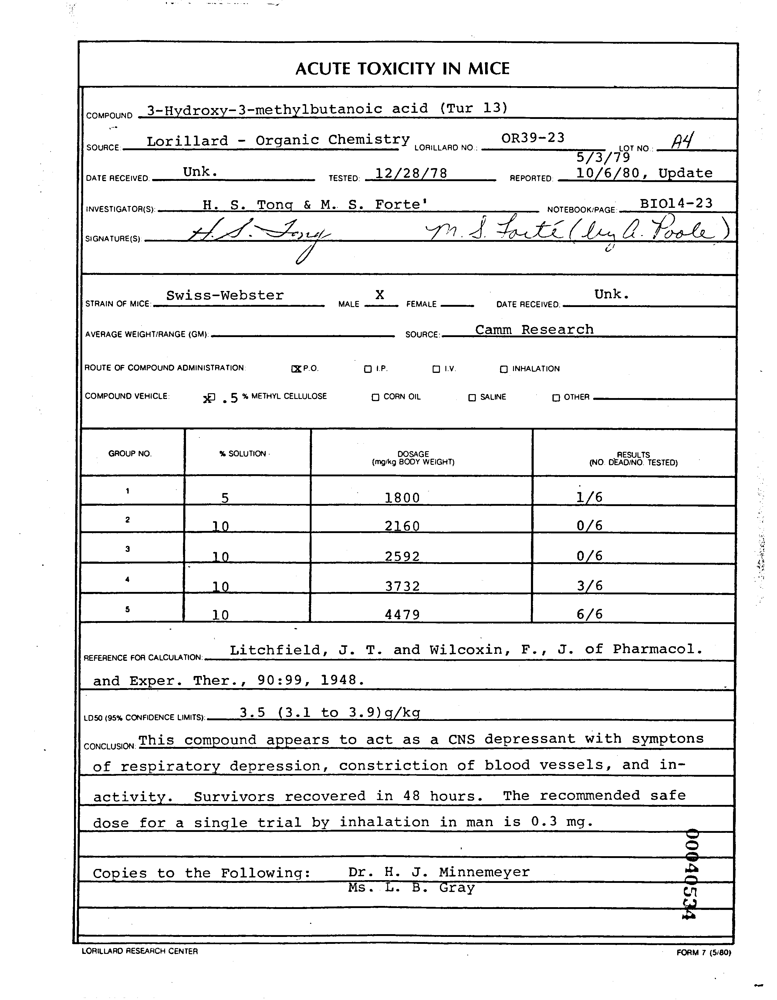}
        \caption{}
        \label{fig:1a}
    \end{subfigure}
    ~ 
    \begin{subfigure}[b]{0.25\textwidth}
        \includegraphics[width=\textwidth]{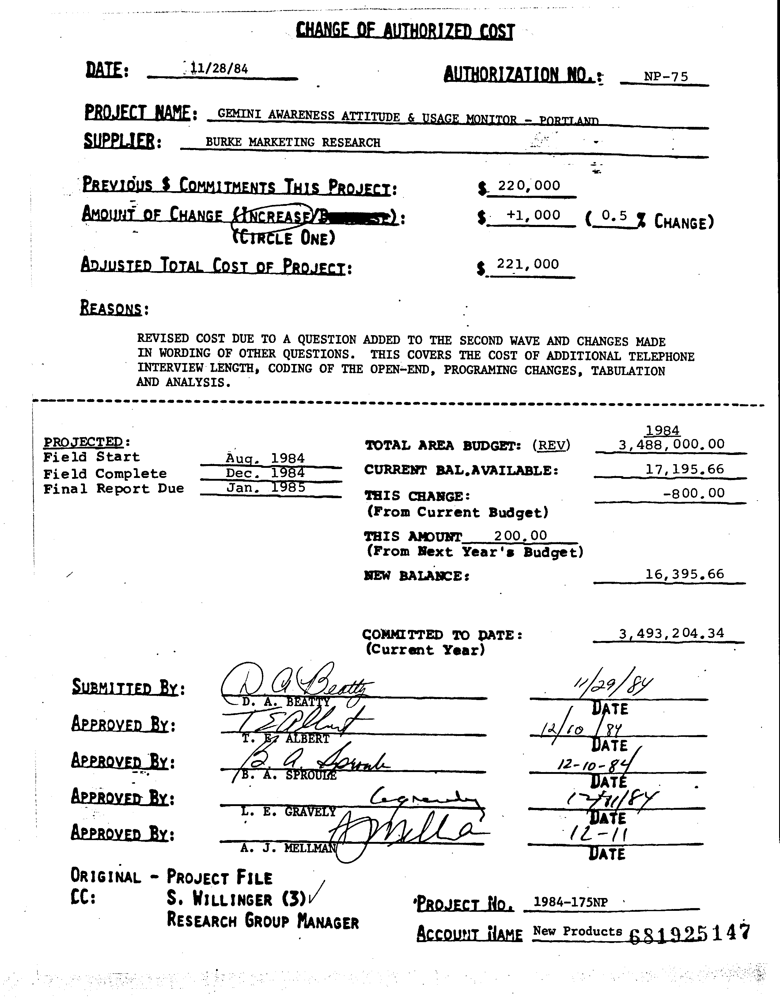}
        \caption{}
        \label{fig:1b}
    \end{subfigure}
    ~ 
    \begin{subfigure}[b]{0.25\textwidth}
        \includegraphics[width=\textwidth]{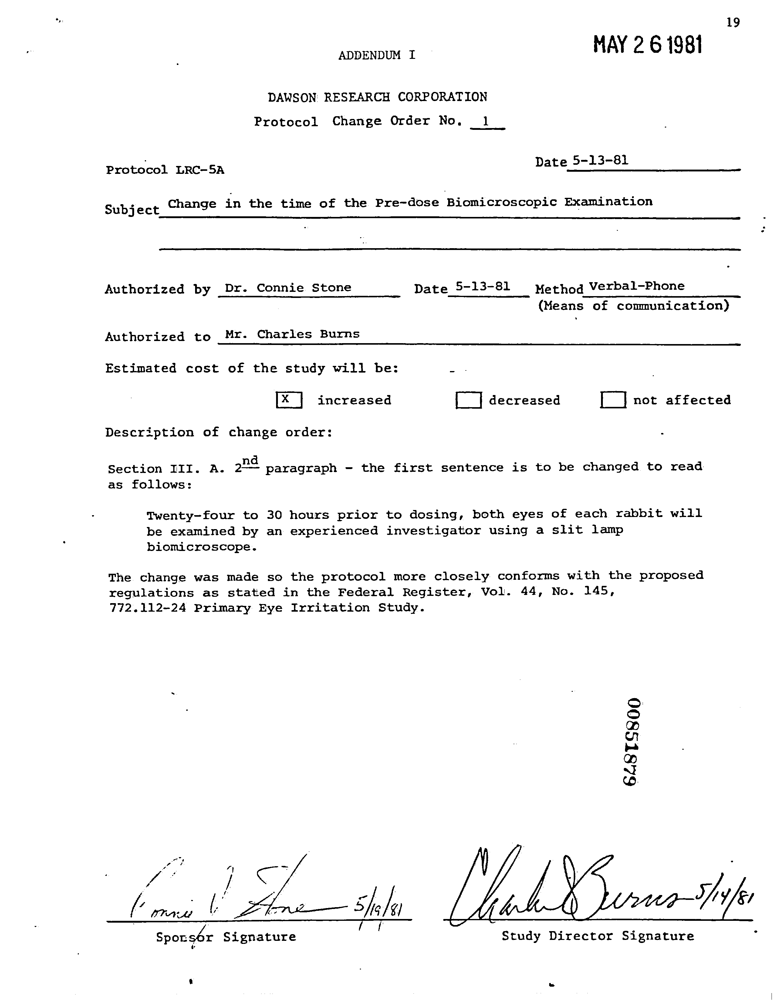}
        \caption{}
        \label{fig:1c}
    \end{subfigure}
    ~
    \begin{subfigure}[b]{0.25\textwidth}
        \includegraphics[width=\textwidth]{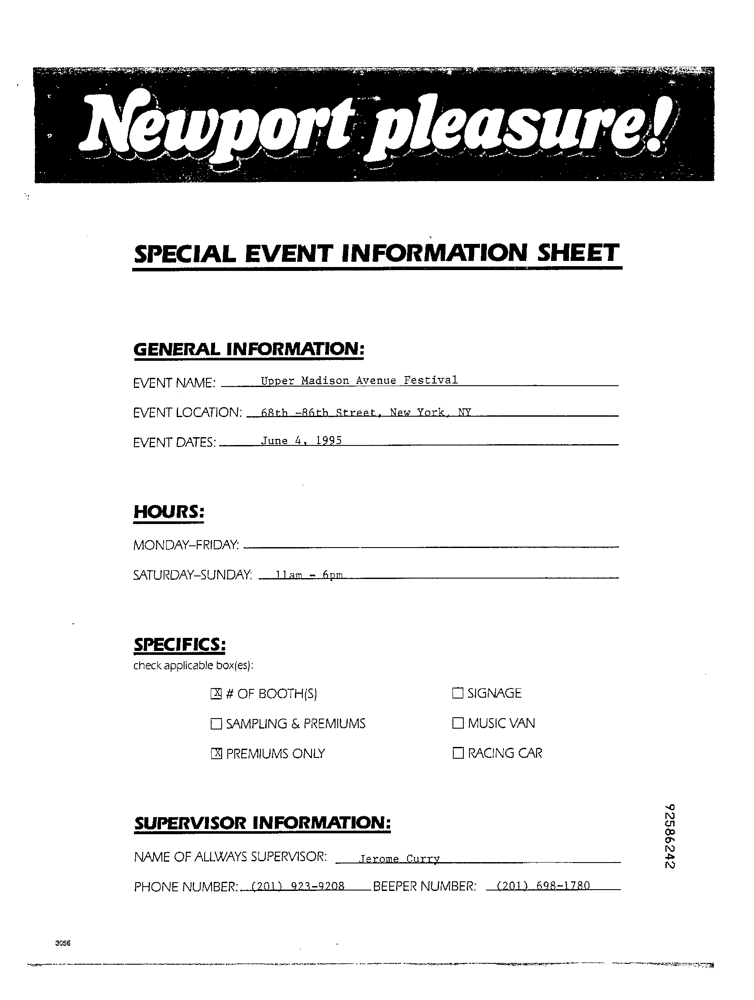}
        \caption{}
        \label{fig:1d}
    \end{subfigure}
    \caption{Scanned images of business documents with different layouts and formats}\label{fig:1}
\end{figure*}

Document AI, or Document Intelligence\footnote{\url{https://sites.google.com/view/di2019}}, is a relatively new research topic that refers techniques for automatically reading, understanding, and analyzing business documents. Business documents are files that provide details related to a company's internal and external transactions, which are shown in Figure~\ref{fig:1}. They may be digital-born, occurring as electronic files, or they may be in scanned form that comes from written or printed on paper. Some common examples of business documents include purchase orders, financial reports, business emails, sales agreements, vendor contracts, letters, invoices, receipts, resumes, and many others. Business documents are critical to a company's efficiency and productivity. The exact format of a business document may vary, but the information is usually presented in natural language and can be organized in a variety of ways from plain text, multi-column layouts, and a wide variety of tables/forms/figures. Understanding business documents is a very challenging task due to the diversity of layouts and formats, poor quality of scanned document images as well as the complexity of template structures.

Nowadays, many companies extract data from business documents through manual efforts that are time-consuming and expensive, meanwhile requiring manual customization or configuration. Rules and workflows for each type of document often need to be hard-coded and updated with changes to the specific format or when dealing with multiple formats. To address these problems, document AI models and algorithms are designed to automatically classify, extract, and structuralize information from business documents, accelerating automated document processing workflows. Contemporary approaches for document AI are usually built upon deep neural networks from a computer vision perspective or a natural language processing perspective, or a combination of them. Early attempts usually focused on detecting and analyzing certain parts of a document, such as tabular areas.~\cite{Hao2016ATD} were the first to propose a table detection method for PDF documents based on Convolutional Neural Networks (CNN). After that, ~\citep{Schreiber2017DeepDeSRTDL,soto-yoo-2019-visual,Zhong2019PubLayNetLD} also leveraged more advanced Faster R-CNN model~\citep{Ren2015FasterRT} or Mask R-CNN model~\citep{DBLP:journals/corr/HeGDG17} to further improve the accuracy of document layout analysis. In addition,~\cite{Yang2017LearningTE} presented an end-to-end, multimodal, fully convolutional network for extracting semantic structures from document images, taking advantage of text embeddings from pre-trained NLP models. More recently,~\cite{liu-etal-2019-graph} introduced a Graph Convolutional Networks (GCN) based model to combine textual and visual information for information extraction from business documents. Although these models have made significant progress in the document AI area with deep neural networks, most of these methods confront two limitations: (1) They rely on a few human-labeled training samples without fully exploring the possibility of using large-scale unlabeled training samples. (2) They usually leverage either pre-trained CV models or NLP models, but do not consider a joint training of textual and layout information. Therefore, it is important to investigate how self-supervised pre-training of text and layout may help in the document AI area.

\begin{figure*}[t]
    \centering
    \includegraphics[width=0.95\textwidth]{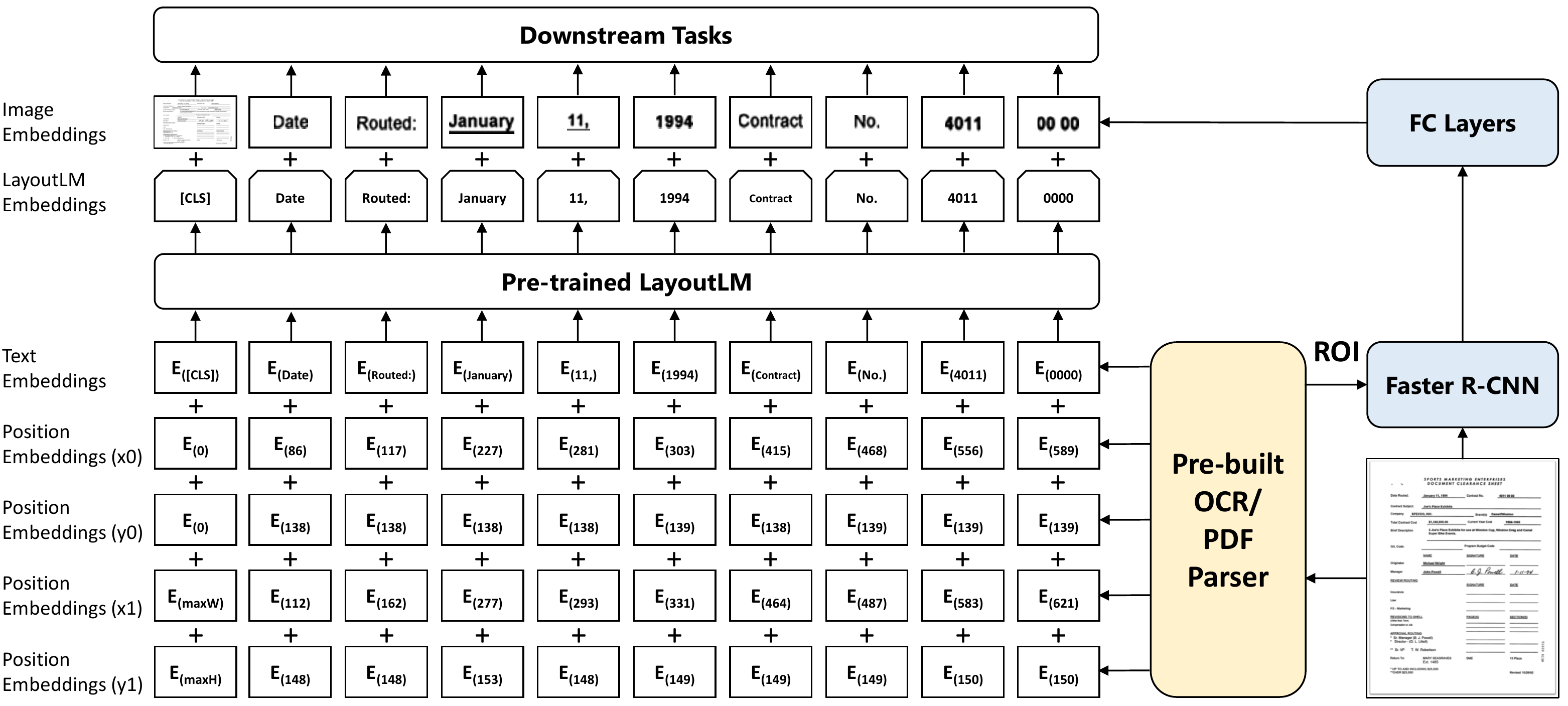}
    \caption{An example of LayoutLM, where 2-D layout and image embeddings are integrated into the original BERT architecture. The LayoutLM embeddings and image embeddings from Faster R-CNN work together for downstream tasks.}
    \label{fig:2}
\end{figure*}

To this end, we propose LayoutLM, a simple yet effective pre-training method of text and layout for document image understanding tasks. Inspired by the BERT model~\citep{devlin-etal-2019-bert}, where input textual information is mainly represented by text embeddings and position embeddings, LayoutLM further adds two types of input embeddings: (1) a 2-D position embedding that denotes the relative position of a token within a document; (2) an image embedding for scanned token images within a document. The architecture of LayoutLM is shown in Figure~\ref{fig:2}. We add these two input embeddings because the 2-D position embedding can capture the relationship among tokens within a document, meanwhile the image embedding can capture some appearance features such as font directions, types, and colors. In addition, we adopt a multi-task learning objective for LayoutLM, including a Masked Visual-Language Model (MVLM) loss and a Multi-label Document Classification (MDC) loss, which further enforces joint pre-training for text and layout. In this work, our focus is the document pre-training based on scanned document images, while digital-born documents are less challenging because they can be considered as a special case where OCR is not required, thus they are out of the scope of this paper. Specifically, the LayoutLM is pre-trained on the IIT-CDIP Test Collection 1.0\footnote{\url{https://ir.nist.gov/cdip/}}~\citep{Lewis:2006:BTC:1148170.1148307}, which contains more than 6 million scanned documents with 11 million scanned document images. The scanned documents are in a variety of categories, including letter, memo, email, filefolder, form, handwritten, invoice, advertisement, budget, news articles, presentation, scientific publication, questionnaire, resume, scientific report, specification, and many others, which is ideal for large-scale self-supervised pre-training. We select three benchmark datasets as the downstream tasks to evaluate the performance of the pre-trained LayoutLM model. 
The first is the FUNSD dataset\footnote{\url{https://guillaumejaume.github.io/FUNSD/}}~\citep{Jaume2019FUNSDAD} that is used for spatial layout analysis and form understanding. The second is the SROIE dataset\footnote{\url{https://rrc.cvc.uab.es/?ch=13}} for Scanned Receipts Information Extraction. The third is the RVL-CDIP dataset\footnote{\url{https://www.cs.cmu.edu/~aharley/rvl-cdip/}}~\citep{Harley2015EvaluationOD} for document image classification, which consists of 400,000 grayscale images in 16 classes. 
Experiments illustrate that the pre-trained LayoutLM model significantly outperforms several SOTA pre-trained models on these benchmark datasets, demonstrating the enormous advantage for pre-training of text and layout information in document image understanding tasks.

The contributions of this paper are summarized as follows:
\begin{itemize}
    \item For the first time, textual and layout information from scanned document images is pre-trained in a single framework. Image features are also leveraged to achieve new state-of-the-art results. 
    \item LayoutLM uses the masked visual-language model and the multi-label document classification as the training objectives, which significantly outperforms several SOTA pre-trained models in document image understanding tasks.
    \item The code and pre-trained models are publicly available at \url{https://aka.ms/layoutlm} for more downstream tasks.
\end{itemize}
 
\section{LayoutLM}
In this section, we briefly review the BERT model, and introduce how we extend to jointly model text and layout information in the LayoutLM framework.

\subsection{The BERT Model}



The BERT model is an attention-based bidirectional language modeling approach. It has been verified that the BERT model shows effective knowledge transfer from the self-supervised task with large-scale training data.
The architecture of BERT is basically a multi-layer bidirectional Transformer encoder. It accepts a sequence of tokens and stacks multiple layers to produce final representations. In detail, given a set of tokens processed using WordPiece, the input embeddings are computed by summing the corresponding word embeddings, position embeddings, and segment embeddings. Then, these input embeddings are passed through a multi-layer bidirectional Transformer that can generate contextualized representations with an adaptive attention mechanism.

There are two steps in the BERT framework: pre-training and fine-tuning. During the pre-training, the model uses two objectives to learn the language representation: Masked Language Modeling (MLM) and Next Sentence Prediction (NSP), where MLM randomly masks some input tokens and the objective is to recover these masked tokens, and NSP is a binary classification task taking a pair of sentences as inputs and classifying whether they are two consecutive sentences. In the fine-tuning, task-specific datasets are used to update all parameters in an end-to-end way. The BERT model has been successfully applied in a set of NLP tasks.

\subsection{The LayoutLM Model}

Although BERT-like models become the state-of-the-art techniques on several challenging NLP tasks, they usually leverage text information only for any kind of inputs. When it comes to visually rich documents, there is much more information that can be encoded into the pre-trained model. Therefore, we propose to utilize the visually rich information from document layouts and align them with the input texts. Basically, there are two types of features which substantially improve the language representation in a visually rich document, which are:

\paragraph{Document Layout Information}
It is evident that the relative positions of words in a document contribute a lot to the semantic representation. Taking form understanding as an example, given a key in a form (e.g., ``Passport ID:''), its corresponding value is much more likely on its right or below instead of on the left or above. Therefore, we can embed these relative positions information as 2-D position representation. Based on the self-attention mechanism within the Transformer, embedding 2-D position features into the language representation will better align the layout information with the semantic representation.


\paragraph{Visual Information}
Compared with the text information, the visual information is another significantly important feature in document representations. Typically, documents contain some visual signals to show the importance and priority of document segments. The visual information can be represented by image features and effectively utilized in document representations. For document-level visual features, the whole image can indicate the document layout, which is an essential feature for document image classification. 
For word-level visual features, styles such as bold, underline, and italic, are also significant hints for the sequence labeling tasks. 
Therefore, we believe that combining the image features with traditional text representations can bring richer semantic representations to documents.

\subsection{Model Architecture}
To take advantage of existing pre-trained models and adapt to document image understanding tasks, we use the BERT architecture as the backbone and add two new input embeddings: a 2-D position embedding and an image embedding.

\paragraph{2-D Position Embedding}
Unlike the position embedding that models the word position in a sequence, 2-D position embedding aims to model the relative spatial position in a document. To represent the spatial position of elements in scanned document images, we consider a document page as a coordinate system with the top-left origin. In this setting, the bounding box can be precisely defined by ($x_0$, $y_0$, $x_1$, $y_1$), where ($x_0$, $y_0$) corresponds to the position of the upper left in the bounding box, and ($x_1$, $y_1$) represents the position of the lower right. We add four position embedding layers with two embedding tables, where the embedding layers representing the same dimension share the same embedding table. This means that we look up the position embedding of $x_0$ and $x_1$ in the embedding table $X$ and lookup $y_0$ and $y_1$ in table $Y$.

\paragraph{Image Embedding} 
To utilize the image feature of a document and align the image feature with the text, we add an image embedding layer to represent image features in language representation. In more detail, with the bounding box of each word from OCR results, we split the image into several pieces, and they have a one-to-one correspondence with the words. We generate the image region features with these pieces of images from the Faster R-CNN~\citep{Ren2015FasterRT} model as the token image embeddings. For the {\tt[CLS]} token, we also use the Faster R-CNN model to produce embeddings using the whole scanned document image as the Region of Interest (ROI) to benefit the downstream tasks which need the representation of the {\tt[CLS]} token.

\subsection{Pre-training LayoutLM}

\paragraph{Task \#1: Masked Visual-Language Model}
Inspired by the masked language model, we propose the Masked Visual-language Model (MVLM) to learn the language representation with the clues of 2-D position embeddings and text embeddings. During the pre-training, we randomly mask some of the input tokens but keep the corresponding 2-D position embeddings, and then the model is trained to predict the masked tokens given the contexts. In this way, the LayoutLM model not only understands the language contexts but also utilizes the corresponding 2-D position information, thereby bridging the gap between the visual and language modalities.

\paragraph{Task \#2: Multi-label Document Classification}
For document image understanding, many tasks require the model to generate high-quality document-level representations. As the IIT-CDIP Test Collection includes multiple tags for each document image, we also use a Multi-label Document Classification (MDC) loss during the pre-training phase. Given a set of scanned documents, we use the document tags to supervise the pre-training process so that the model can cluster the knowledge from different domains and generate better document-level representation. Since the MDC loss needs the label for each document image that may not exist for larger datasets, it is optional during the pre-training and may not be used for pre-training larger models in the future. We will compare the performance of MVLM and MVLM+MDC in Section 3.

\subsection{Fine-tuning LayoutLM}
The pre-trained LayoutLM model is fine-tuned on three document image understanding tasks, including a form understanding task, a receipt understanding task as well as a document image classification task. For the form and receipt understanding tasks, LayoutLM predicts \{B, I, E, S, O\} tags for each token and uses sequential labeling to detect each type of entity in the dataset. For the document image classification task, LayoutLM predicts the class labels using the representation of the {\tt[CLS]} token.

\section{Experiments}
\subsection{Pre-training Dataset}
The performance of pre-trained models is largely determined by the scale and quality of datasets. Therefore, we need a large-scale scanned document image dataset to pre-train the LayoutLM model. Our model is pre-trained on the IIT-CDIP Test Collection 1.0, which contains more than 6 million documents, with more than 11 million scanned document images. Moreover, each document has its corresponding text and metadata stored in XML files. The text is the content produced by applying OCR to document images. The metadata describes the properties of the document, such as the unique identity and document labels. Although the metadata contains erroneous and inconsistent tags, the scanned document images in this large-scale dataset are perfectly suitable for pre-training our model.

\subsection{Fine-tuning Dataset}

\paragraph{The FUNSD Dataset} 
We evaluate our approach on the FUNSD dataset for form understanding in noisy scanned documents. This dataset includes 199 real, fully annotated, scanned forms with 9,707 semantic entities and 31,485 words. These forms are organized as a list of semantic entities that are interlinked. Each semantic entity comprises a unique identifier, a label (i.e., question, answer, header, or other), a bounding box, a list of links with other entities, and a list of words. The dataset is split into 149 training samples and 50 testing samples. We adopt the word-level F1 score as the evaluation metric.

\paragraph{The SROIE Dataset}

We also evaluate our model on the SROIE dataset for receipt information extraction (Task 3). The dataset contains 626 receipts for training and 347 receipts for testing. Each receipt is organized as a list of text lines with bounding boxes. Each receipt is labeled with four types of entities which are \{company, date, address, total\}. The evaluation metric is the exact match of the entity recognition results in the F1 score.

\paragraph{The RVL-CDIP Dataset}

The RVL-CDIP dataset consists of 400,000 grayscale images in 16 classes, with 25,000 images per class. There are 320,000 training images, 40,000 validation images, and 40,000 test images. The images are resized, so their largest dimension does not exceed 1,000 pixels. The 16 classes include \{letter, form, email, handwritten, advertisement, scientific report, scientific publication, specification, file folder, news article, budget, invoice, presentation, questionnaire, resume, memo\}. The evaluation metric is the overall classification accuracy.

\subsection{Document Pre-processing}
To utilize the layout information of each document, we need to obtain the location of each token. However, the pre-training dataset (IIT-CDIP Test Collection) only contains pure texts while missing their corresponding bounding boxes. In this case, we re-process the scanned document images to obtain the necessary layout information. Like the original pre-processing in IIT-CDIP Test Collection, we similarly process the dataset by applying OCR to document images. The difference is that we obtain both the recognized words and their corresponding locations in the document image. Thanks to Tesseract\footnote{\url{https://github.com/tesseract-ocr/tesseract}}, an open-source OCR engine, we can easily obtain the recognition as well as the 2-D positions. We store the OCR results in hOCR format, a standard specification format which clearly defines the OCR results of one single document image using a hierarchical representation.



\subsection{Model Pre-training}
We initialize the weight of LayoutLM model with the pre-trained BERT base model. Specifically, our BASE model has the same architecture: a 12-layer Transformer with 768 hidden sizes, and 12 attention heads, which contains about 113M parameters. Therefore, we use the BERT base model to initialize all modules in our model except the 2-D position embedding layer. For the LARGE setting, our model has a 24-layer Transformer with 1,024 hidden sizes and 16 attention heads, which is initialized by the pre-trained BERT LARGE model and contains about 343M parameters. Following ~\citep{devlin-etal-2019-bert}, we select 15\% of the input tokens for prediction. We replace these masked tokens with the {\tt[MASK]} token 80\% of the time, a random token 10\% of the time, and an unchanged token 10\% of the time. Then, the model predicts the corresponding token with the cross-entropy loss.

In addition, we also add the 2-D position embedding layers with four embedding representations ($x_0$, $y_0$, $x_1$, $y_1$), where ($x_0$, $y_0$) corresponds to the position of the upper left in the bounding box, and ($x_1$, $y_1$) represents the position of the lower right. Considering that the document layout may vary in different page size, we scale the actual coordinate to a ``virtual'' coordinate: the actual coordinate is scaled to have a value from 0 to 1,000. Furthermore, we also use the ResNet-101 model as the backbone network in the Faster R-CNN model, which is pre-trained on the Visual Genome dataset~\citep{krishnavisualgenome}.

We train our model on 8 NVIDIA Tesla V100 32GB GPUs with a total batch size of 80. The Adam optimizer is used with an initial learning rate of 5e-5 and a linear decay learning rate schedule. The BASE model takes 80 hours to finish one epoch on 11M documents, while the LARGE model takes nearly 170 hours to finish one epoch.

\subsection{Task-specific Fine-tuning}
We evaluate the LayoutLM model on three document image understanding tasks: \textbf{Form Understanding}, \textbf{Receipt Understanding}, and \textbf{Document Image Classification}. 
We follow the typical fine-tuning strategy and update all parameters in an end-to-end way on task-specific datasets.

\paragraph{Form Understanding}
This task requires extracting and structuring the textual content of forms. It aims to extract key-value pairs from the scanned form images. In more detail, this task includes two sub-tasks: semantic labeling and semantic linking. Semantic labeling is the task of aggregating words as semantic entities and assigning pre-defined labels to them. Semantic linking is the task of predicting the relations between semantic entities. In this work, we focus on the semantic labeling task, while semantic linking is out of the scope.  To fine-tune LayoutLM on this task, we treat semantic labeling as a sequence labeling problem. We pass the final representation into a linear layer followed by a softmax layer to predict the label of each token. The model is trained for 100 epochs with a batch size of 16 and a learning rate of 5e-5.

\paragraph{Receipt Understanding}
This task requires filling several pre-defined semantic slots according to the scanned receipt images. For instance, given a set of receipts, we need to fill specific slots ( i.g., company, address, date, and total). Different from the form understanding task that requires labeling all matched entities and key-value pairs, the number of semantic slots is fixed with pre-defined keys. Therefore, the model only needs to predict the corresponding values using the sequence labeling method.

\paragraph{Document Image Classification}
Given a visually rich document, this task aims to predict the corresponding category for each document image. Distinct from the existing image-based approaches, our model includes not only image representations but also text and layout information using the multimodal architecture in LayoutLM. Therefore, our model can combine the text, layout, and image information in a more effective way. To fine-tune our model on this task, 
we concatenate the output from the LayoutLM model and the whole image embedding, followed by a softmax layer for category prediction. We fine-tune the model for 30 epochs with a batch size of 40 and a learning rate of 2e-5.

\subsection{Results}

\paragraph{Form Understanding}

We evaluate the form understanding task on the FUNSD dataset. The experiment results are shown in Table~\ref{tab:1}. We compare the LayoutLM model with two SOTA pre-trained NLP models: BERT and RoBERTa~\citep{Liu2019RoBERTaAR}. The BERT BASE model achieves 0.603 and while the LARGE model achieves 0.656 in F1. Compared to BERT, the RoBERTa performs much better on this dataset as it is trained using larger data with more epochs. Due to the time limitation, we present 4 settings for LayoutLM, which are 500K document pages with 6 epochs, 1M with 6 epochs, 2M with 6 epochs as well as 11M with 2 epochs. It is observed that the LayoutLM model substantially outperforms existing SOTA pre-training baselines. With the BASE architecture, the LayoutLM model with 11M training data achieves 0.7866 in F1, which is much higher than BERT and RoBERTa with the similar size of parameters. In addition, we also add the MDC loss in the pre-training step and it does bring substantial improvements on the FUNSD dataset. 
Finally, the LayoutLM model achieves the best performance of 0.7927 when using the text, layout, and image information at the same time.

\begin{table*}[ht]
    \centering
    \begin{tabular}{clcccc}
    \toprule
     \bf Modality &  \multicolumn{1}{c}{\bf Model} & \bf Precision & \bf Recall & \bf F1 & \bf \#Parameters  \\\midrule
     \multirow{4}{*}{Text only}& $\textrm{BERT}_{\rm BASE}$  & 0.5469 & 0.671 & 0.6026 & 110M \\
     & $\textrm{RoBERTa}_{\rm BASE}$  & 0.6349 & 0.6975 & 0.6648 & 125M  \\
     & $\textrm{BERT}_{\rm LARGE}$   & 0.6113 & 0.7085 & 0.6563 & 340M \\
     & $\textrm{RoBERTa}_{\rm LARGE}$  & 0.678 & 0.7391 & 0.7072 & 355M \\\midrule
     \multirow{4}{*}{\makecell{Text + Layout \\MVLM}}& $\textrm{LayoutLM}_{\rm BASE}$~(500K, 6 epochs)   & 0.665 & 0.7355 & 0.6985 & 113M\\
     &$\textrm{LayoutLM}_{\rm BASE}$~(1M, 6 epochs) & 0.6909 & 0.7735 & 0.7299 & 113M \\
     &$\textrm{LayoutLM}_{\rm BASE}$~(2M, 6 epochs)  & 0.7377 & 0.782  & 0.7592 & 113M\\
     &$\textrm{LayoutLM}_{\rm BASE}$~(11M, 2 epochs) & 0.7597 & 0.8155 & 0.7866 & 113M \\\cmidrule{2-6}
     \multirow{2}{*}{\makecell{Text + Layout \\MVLM+MDC}}
     &$\textrm{LayoutLM}_{\rm BASE}$~(1M, 6 epochs) & 0.7076 & 0.7695 & 0.7372 & 113M\\
     &$\textrm{LayoutLM}_{\rm BASE}$~(11M, 1 epoch) & 0.7194 & 0.7780 & 0.7475 & 113M\\\midrule
     \multirow{2}{*}{\makecell{Text + Layout \\MVLM}}& $\textrm{LayoutLM}_{\rm LARGE}$~(1M, 6 epochs)  & 0.7171 & 0.805 & 0.7585 & 343M \\
     & $\textrm{LayoutLM}_{\rm LARGE}$~(11M, 1 epoch) & 0.7536 & 0.806 & 0.7789 & 343M \\\midrule
     \multirow{2}{*}{\makecell{Text + Layout + Image \\MVLM}}& $\textrm{LayoutLM}_{\rm BASE}$~(1M, 6 epochs)  & 0.7101 & 0.7815 & 0.7441 & 160M \\
     & $\textrm{LayoutLM}_{\rm BASE}$~(11M, 2 epochs) & \bf 0.7677 & \bf 0.8195 & \bf 0.7927 & 160M \\
     \bottomrule
    \end{tabular}
    \caption{Model accuracy (Precision, Recall, F1) on the FUNSD dataset}
    \label{tab:1}
\end{table*}

\begin{table*}[ht]
    \centering
    \begin{tabular}{ccccc}
    \toprule
     \bf \# Pre-training Data & \bf \# Pre-training Epochs & \bf Precision & \bf Recall & \bf F1   \\\midrule
     \multirow{6}{*}{500K} & 1 epoch   & 0.5779  & 0.6955 & 0.6313  \\
     & 2 epochs   & 0.6217 & 0.705 & 0.6607  \\
     & 3 epochs   & 0.6304 & 0.718 & 0.6713  \\
     & 4 epochs   & 0.6383 & 0.7175 & 0.6756  \\
     & 5 epochs   & 0.6568 & 0.734 & 0.6933  \\
     & 6 epochs   & 0.665 & 0.7355 & 0.6985  \\\midrule
     \multirow{6}{*}{1M}& 1 epoch   & 0.6156 & 0.7005  & 0.6552  \\
     & 2 epochs  & 0.6545	& 0.737  & 0.6933  \\
     & 3 epochs  & 0.6794	& 0.762 & 0.7184  \\
     & 4 epochs  & 0.6812	& 0.766 & 0.7211  \\
     & 5 epochs  & 0.6863	& 0.7625 & 0.7224  \\
     & 6 epochs  & 0.6909	& 0.7735 & 0.7299  \\\midrule
     \multirow{6}{*}{2M} & 1 epoch   & 0.6599 & 0.7355 & 0.6957 \\
      & 2 epochs   & 0.6938  & 0.759  & 0.7249  \\
      & 3 epochs   & 0.6915 & 0.7655  & 0.7266  \\
      & 4 epochs   & 0.7081 & 0.781  & 0.7427  \\
      & 5 epochs   & 0.7228 & 0.7875  & 0.7538  \\
      & 6 epochs   & 0.7377 & 0.782  & 0.7592  \\\midrule
     \multirow{2}{*}{11M} & 1 epoch  & 0.7464 & 0.7815 & 0.7636  \\
     & 2 epochs  & \bf 0.7597 & \bf 0.8155  & \bf 0.7866 \\
     \bottomrule
    \end{tabular}
    \caption{$\textrm{LayoutLM}_{\rm BASE}$~(Text + Layout, MVLM) accuracy with different data and epochs on the FUNSD dataset}
    \label{tab:2}
\end{table*}

In addition, we also evaluate the LayoutLM model with different data and epochs on the FUNSD dataset, which is shown in Table~\ref{tab:2}. For different data settings, we can see that the overall accuracy is monotonically increased as more epochs are trained during the pre-training step. Furthermore, the accuracy is also improved as more data is fed into the LayoutLM model. As the FUNSD dataset contains only 149 images for fine-tuning, the results confirm that the pre-training of text and layout is effective for scanned document understanding especially with low resource settings.

Furthermore, we compare different initialization methods for the LayoutLM model including from scratch, BERT and RoBERTa. The results in Table~\ref{tab:3} show that the LayoutLM$_{\rm BASE}$ model initialized with RoBERTa$_{\rm BASE}$ outperforms BERT$_{\rm BASE}$ by 2.1 points in F1. For the LARGE setting, the LayoutLM$_{\rm LARGE}$ model initialized with RoBERTa$_{\rm LARGE}$ further improve 1.3 points over the BERT$_{\rm LARGE}$ model. We will pre-train more models with RoBERTa as the initialization in the future, especially for the LARGE settings.

\begin{table*}[ht]
    \centering
    \begin{tabular}{ccccc}
    \toprule
     \bf Initialization &\bf Model & \bf Precision & \bf Recall & \bf F1   \\\midrule
     $\textrm{SCRATCH}$ & $\textrm{LayoutLM}_{\rm BASE}$~(1M, 6 epochs)  & 0.5630  & 0.6728 & 0.6130  \\
     $\textrm{BERT}_{\rm BASE}$ & $\textrm{LayoutLM}_{\rm BASE}$~(1M, 6 epochs)  & 0.6909  & 0.7735 & 0.7299 \\
     $\textrm{RoBERTa}_{\rm BASE}$ & $\textrm{LayoutLM}_{\rm BASE}$~(1M, 6 epochs)   & 0.7173 & 0.7888 & 0.7514 \\
     $\textrm{SCRATCH}$ & $\textrm{LayoutLM}_{\rm LARGE}$~(11M, 1 epoch) & 0.6845 & 0.7804 & 0.7293  \\
     $\textrm{BERT}_{\rm LARGE}$ & $\textrm{LayoutLM}_{\rm LARGE}$~(11M, 1 epoch)  & 0.7536  & 0.8060 & 0.7789 \\
     $\textrm{RoBERTa}_{\rm LARGE}$ & $\textrm{LayoutLM}_{\rm LARGE}$~(11M, 1 epoch)   & 0.7681 & 0.8188 & 0.7926 \\
    \bottomrule
    \end{tabular}
    \caption{Different initialization methods for $\rm BASE$ and $\rm LARGE$~(Text + Layout, MVLM)}
    \label{tab:3}
\end{table*}

\paragraph{Receipt Understanding}

We evaluate the receipt understanding task using the SROIE dataset. The results are shown in Table~\ref{tab:4}. As we only test the performance of the Key Information Extraction task in SROIE, we would like to eliminate the effect of incorrect OCR results. Therefore, we pre-process the training data by using the ground truth OCR and run a set of experiments using the baseline models (BERT \& RoBERTa) as well as the LayoutLM model. The results show that the $\textrm{LayoutLM}_{\rm LARGE}$ model trained with 11M document images achieve an F1 score of 0.9524, which is significantly better than the first place in the competition leaderboard. This result also verifies that the pre-trained LayoutLM not only performs well on the in-domain dataset (FUNSD) but also outperforms several strong baselines on the out-of-domain dataset like SROIE.

\begin{table*}[ht]
    \centering
    \begin{tabular}{clcccc}
    \toprule
     \bf Modality &  \multicolumn{1}{c}{\bf Model} & \bf Precision & \bf Recall & \bf F1   & \bf \#Parameters\\\midrule
     \multirow{4}{*}{Text only}&$\textrm{BERT}_{\rm BASE}$  & 0.9099 & 0.9099 & 0.9099 & 110M\\
     &$\textrm{RoBERTa}_{\rm BASE}$  & 0.9107 & 0.9107 & 0.9107 &  125M\\
     &$\textrm{BERT}_{\rm LARGE}$   & 0.9200  & 0.9200  & 0.9200 &  340M\\
     &$\textrm{RoBERTa}_{\rm LARGE}$  & 0.9280 & 0.9280  & 0.9280 & 355M\\\midrule
     \multirow{4}{*}{\makecell{Text + Layout\\MVLM}}
     &$\textrm{LayoutLM}_{\rm BASE}$~(500K, 6 epochs) & 0.9388 & 0.9388& 0.9388 & 113M\\
     &$\textrm{LayoutLM}_{\rm BASE}$~(1M, 6 epochs) &	0.9380 & 0.9380& 0.9380 & 113M\\
     &$\textrm{LayoutLM}_{\rm BASE}$~(2M, 6 epochs)   & 0.9431 & 0.9431& 0.9431 & 113M\\
     &$\textrm{LayoutLM}_{\rm BASE}$~(11M, 2 epochs)  & 0.9438 & 0.9438 & 0.9438 & 113M \\\cmidrule{2-6}
     \multirow{2}{*}{\makecell{Text + Layout \\MVLM+MDC}}&$\textrm{LayoutLM}_{\rm BASE}$~(1M, 6 epochs) & 0.9402& 0.9402& 0.9402 & 113M\\
     &$\textrm{LayoutLM}_{\rm BASE}$~(11M, 1 epoch) & 0.9460 & 0.9460& 0.9460  & 113M\\
     \midrule
     
     \multirow{2}{*}{\makecell{Text + Layout \\MVLM}}& $\textrm{LayoutLM}_{\rm LARGE}$~(1M, 6 epochs)  & 0.9416 & 0.9416 & 0.9416 & 343M \\
     & $\textrm{LayoutLM}_{\rm LARGE}$~(11M, 1 epoch) & \bf 0.9524 & \bf 0.9524 & \bf 0.9524 & 343M \\\midrule
     \multirow{2}{*}{\makecell{Text + Layout + Image \\MVLM}}& $\textrm{LayoutLM}_{\rm BASE}$~(1M, 6 epochs)  & 0.9416 & 0.9416 & 0.9416 & 160M\\
     & $\textrm{LayoutLM}_{\rm BASE}$~(11M, 2 epochs) & 0.9467 & 0.9467 & 0.9467 & 160M\\\midrule
     
     Baseline & Ranking $1^{st}$ in SROIE & 0.9402 & 0.9402 & 0.9402 & -\\
     \bottomrule
    \end{tabular}
    \caption{Model accuracy (Precision, Recall, F1) on the SROIE dataset}
    \label{tab:4}
\end{table*}

\paragraph{Document Image Classification}

Finally, we evaluate the document image classification task using the RVL-CDIP dataset. Document images are different from other natural images as most of the content in document images are texts in a variety of styles and layouts. Traditionally, image-based classification models with pre-training perform much better than the text-based models, which is shown in Table~\ref{tab:5}. We can see that either BERT or RoBERTa underperforms the image-based approaches, illustrating that text information is not sufficient for this task, and it still needs layout and image features. We address this issue by using the LayoutLM model for this task. Results show that, even without the image features, LayoutLM still outperforms the single model of the image-based approaches. 
After integrating the image embeddings, the LayoutLM achieves the accuracy of 94.42\%, which is significantly better than several SOTA baselines for document image classification. It is observed that our model performs best in the "email" category while performs worst in the "form" category. We will further investigate how to take advantage of both pre-trained LayoutLM and image models, as well as involve image information in the pre-training step for the LayoutLM model.

\begin{table*}[t]
    \centering
    \begin{tabular}{clcc}
    \toprule
     \bf Modality & \multicolumn{1}{c}{\bf Model} & \bf Accuracy   & \bf \#Parameters\\\midrule
     \multirow{4}{*}{Text only} &$\textrm{BERT}_{\rm BASE}$  &  89.81\% & 110M\\
     &$\textrm{RoBERTa}_{\rm BASE}$  &  90.06\%    & 125M\\
     &$\textrm{BERT}_{\rm LARGE}$   & 89.92\%    & 340M\\
     &$\textrm{RoBERTa}_{\rm LARGE}$  & 90.11\%   &  355M\\\midrule
      \multirow{4}{*}{\makecell{Text + Layout \\MVLM}}
      &$\textrm{LayoutLM}_{\rm BASE}$~(500K, 6 epochs)   & 91.25\%   & 113M\\
     &$\textrm{LayoutLM}_{\rm BASE}$~(1M, 6 epochs)  &  91.48\%   & 113M\\
     &$\textrm{LayoutLM}_{\rm BASE}$~(2M, 6 epochs)  &  91.65\%   & 113M\\
     &$\textrm{LayoutLM}_{\rm BASE}$~(11M, 2 epochs) &  91.78\% & 113M\\\cmidrule{2-4}
     \multirow{2}{*}{\makecell{Text + Layout \\MVLM+MDC}}&$\textrm{LayoutLM}_{\rm BASE}$~(1M, 6 epochs) & 91.74\% & 113M\\
     &$\textrm{LayoutLM}_{\rm BASE}$~(11M, 1 epoch) &  91.78\% & 113M\\
     \midrule
     \multirow{2}{*}{\makecell{Text + Layout \\MVLM}}& $\textrm{LayoutLM}_{\rm LARGE}$~(1M, 6 epochs)  & 91.88\% & 343M \\
     & $\textrm{LayoutLM}_{\rm LARGE}$~(11M, 1 epoch) & 91.90\% & 343M\\
     \midrule
     \multirow{2}{*}{\makecell{Text + Layout + Image \\MVLM}}& $\textrm{LayoutLM}_{\rm BASE}$~(1M, 6 epochs)  & 94.31\% & 160M \\
     & $\textrm{LayoutLM}_{\rm BASE}$~(11M, 2 epochs) & \bf 94.42\% & 160M \\\midrule\midrule
     
     \multirow{7}{*}{\makecell{Baselines}} & VGG-16~\citep{Afzal2017CuttingTE} & 90.97\% &- \\
     & Stacked CNN Single~\citep{Das2018DocumentIC} & 91.11\% & -\\ 
      & Stacked CNN Ensemble~\citep{Das2018DocumentIC} & 92.21\% & -\\ 
      & InceptionResNetV2~\citep{Szegedy2016Inceptionv4IA} & 92.63\% & -\\
      & LadderNet~\citep{ijcai2019-466} & 92.77\% & -\\
      & Multimodal Single~\citep{Dauphinee2019ModularMA} &  93.03\% & -\\ 
      & Multimodal Ensemble~\citep{Dauphinee2019ModularMA} &  93.07\% & -\\ 
     \bottomrule
    \end{tabular}
    \caption{Classification accuracy on the RVL-CDIP dataset}
    \label{tab:5}
\end{table*}



\section{Related Work}

The research of Document Analysis and Recognition (DAR) dates to the early 1990s. The mainstream approaches can be divided into three categories: rule-based approaches, conventional machine learning approaches, and deep learning approaches.

\subsection{Rule-based Approaches}

The rule-based approaches~\citep{lebourgeois1992fast, 244677, ha1995recursive, simon1997fast} contain two types of analysis methods: bottom-up and top-down. Bottom-up methods~\citep{lebourgeois1992fast, ha1995document, simon1997fast} usually detect the connected components of black pixels as the basic computational units in document images, and the document segmentation process is to combine them into higher-level structures through different heuristics and label them according to different structural features. Docstrum algorithm~\citep{244677} is among the earliest successful bottom-up algorithms that are based on the connected component analysis. It groups connected components on a polar structure to derive the final segmentation. \cite{simon1997fast} use a special distance-metric between different components to construct a physical page structure. They further reduced the time complexity by using heuristics and path compression algorithms.

The top-down methods often recursively split a page into columns, blocks, text lines, and tokens. \cite{ha1995recursive} propose replacing the basic unit with the black pixels from all the pixels, and the method decomposed the document using the recursive the X-Y cut algorithm to establish an X-Y tree, which makes complex documents decompose more easily. Although these methods perform well on some documents, they require extensive human efforts to figure out better rules, while sometimes failing to generalize to documents from other sources. Therefore, it is inevitable to leverage machine learning approaches in the DAR research.

\subsection{Machine Learning Approaches}

With the development of conventional machine learning, statistical machine learning approaches~\citep{shilman2005learning,1359749} have become the mainstream for document segmentation tasks during the past decade.~\cite{shilman2005learning} consider the layout information of a document as a parsing problem, and globally search the optimal parsing tree based on a grammar-based loss function. They utilize a machine learning approach to select features and train all parameters during the parsing process. Meanwhile, artificial neural networks~\citep{1359749} have been extensively applied to document analysis and recognition. Most efforts have been devoted to the recognition of isolated handwritten and printed characters with widely recognized successful results. In addition to the ANN model, SVM and GMM~\citep{6628808} have been used in document layout analysis tasks. For machine learning approaches, they are usually time-consuming to design manually crafted features and difficult to obtain a highly abstract semantic context. In addition, these methods usually relied on visual cues but ignored textual information.

\subsection{Deep Learning Approaches}

Recently, deep learning methods have become the mainstream and de facto standard for many machine learning problems. Theoretically, they can fit any arbitrary functions through the stacking of multi-layer neural networks and have been verified to be effective in many research areas.~\cite{Yang2017LearningTE} treat the document semantic structure extraction task as a pixel-by-pixel classification problem. They propose a multimodal neural network that considers visual and textual information, while the limitation of this work is that they only used the network to assist heuristic algorithms to classify candidate bounding boxes rather than an end-to-end approach.~\cite{Viana2017FastCD} propose a lightweight model of document layout analysis for mobile and cloud services. The model uses one-dimensional information of images for inference and compares it with the model using two-dimensional information, achieving comparable accuracy in the experiments.~\cite{katti-etal-2018-chargrid} make use of a fully convolutional encoder-decoder network that predicts a segmentation mask and bounding boxes, and the model significantly outperforms approaches based on sequential text or document images.~\cite{soto-yoo-2019-visual} incorporate contextual information into the Faster R-CNN model that involves the inherently localized nature of article contents to improve region detection performance.

Existing deep learning approaches for DAR usually confront two limitations: (1) The models often rely on limited labeled data while leaving a large amount of unlabeled data unused. (2) Current deep learning models usually leverage pre-trained CV models or NLP models, but do not consider the joint pre-training of text and layout. LayoutLM addresses these two limitations and achieves much better performance compared with the previous baselines.

\section{Conclusion and Future Work}

We present LayoutLM, a simple yet effective pre-training technique with text and layout information in a single framework. Based on the Transformer architecture as the backbone, LayoutLM takes advantage of multimodal inputs, including token embeddings, layout embeddings, and image embeddings. Meanwhile, the model can be easily trained in a self-supervised way based on large scale unlabeled scanned document images. We evaluate the LayoutLM model on three tasks: form understanding, receipt understanding, and scanned document image classification. Experiments show that LayoutLM substantially outperforms several SOTA pre-trained models in these tasks.

For future research, we will investigate pre-training models with more data and more computation resources. In addition, we will also train LayoutLM using the LARGE architecture with text and layout, as well as involving image embeddings in the pre-training step. Furthermore, we will explore new network architectures and other self-supervised training objectives that may further unlock the power of LayoutLM.

\bibliographystyle{ACM-Reference-Format}
\bibliography{sample-base}










\end{document}